\documentclass[10pt,twocolumn,letterpaper]{article}

\usepackage[pagenumbers]{cvpr} 

\usepackage{tikz}
\usetikzlibrary{shapes,arrows,positioning,calc,fit,backgrounds}
\usepackage{graphicx}
\usepackage{subcaption}

\definecolor{cvprblue}{rgb}{0.21,0.49,0.74}
\usepackage[pagebackref,breaklinks,colorlinks,allcolors=cvprblue]{hyperref}

\def\paperID{14} 
\def\confName{CVPR}
\def\confYear{2026}

\title{Unlocking Multi-Site Clinical Data: A Federated Approach to Privacy-First Child Autism Behavior Analysis}

\author{
Guangyu Sun\textsuperscript{1},
Wenhan Wu\textsuperscript{3},
Zhishuai Guo\textsuperscript{4},
Ziteng Wang\textsuperscript{5},
Pegah Khosravi\textsuperscript{1, 2},
Chen Chen\textsuperscript{1}
\\
\textsuperscript{1} Institute of Artificial Intelligence, University of Central Florida \\
\textsuperscript{2} Department of Clinical Sciences, College of Medicine, University of Central Florida\\
\textsuperscript{3} Department of Computer Science, University of North Carolina at Charlotte \\
\textsuperscript{4} Department of Computer Science, Northern Illinois University \\
\textsuperscript{5} Industrial and Systems Engineering, Northern Illinois University \\
\small{\tt{guangyu@ucf.edu, wwu25@charlotte.edu, zguo@niu.edu, zwang3@niu.edu,}} \\
\small{\tt{\{pegah.khosravi, chen.chen\}@ucf.edu}}
}

\begin{document}
\maketitle
\begin{abstract}
Automated recognition of autistic behaviors in children is essential for early intervention and objective clinical assessment. However, the development of robust models is severely hindered by strict privacy regulations (e.g., HIPAA) and the sensitive nature of pediatric data, which prevents the centralized aggregation of clinical datasets. Furthermore, individual clinical sites often suffer from data scarcity, making it difficult to learn generalized behavior patterns or tailor models to site-specific patient distributions. To address these challenges, we observe that Federated Learning (FL) can decouple model training from raw data access, enabling multi-site collaboration while maintaining strict data residency. In this paper, we present the first study exploring Federated Learning for pose-based child autism behavior recognition. Our framework employs a two-layer privacy protection mechanism: utilizing human skeletal abstraction to remove identifiable visual information from the raw RGB videos and FL to ensure sensitive pose data remains within the clinic. This approach leverages distributed clinical data to learn generalized representations while providing the flexibility for site-specific personalization. Experimental results on the MMASD benchmark demonstrate that our framework achieves high recognition accuracy, outperforming traditional federated baselines and providing a robust, privacy-first solution for multi-site clinical analysis.
\end{abstract}
    
\section{Introduction}
\label{sec:intro}

Automated recognition of autism-related behaviors in children has emerged as a transformative tool for early screening, clinical diagnosis, and longitudinal monitoring \cite{barami2024automated, perochon2023early, shamhan2025advancements, said2024imitasd,deng2024language}. Traditionally, the clinical assessment of Autism Spectrum Disorder (ASD) has relied on standardized observational instruments such as the Autism Diagnostic Observation Schedule (ADOS) and the Modified Checklist for Autism in Toddlers (M-CHAT) \cite{barami2024automated}. While these tools are clinically validated, they are inherently limited by human observer bias, significant inter-rater variability, and the substantial resource burden required for administration. Expert clinicians must often observe hours of behavioral interactions to identify subtle stereotypical motor movements or social-communicative deficits. By leveraging computer vision to quantify repetitive or atypical movements, automated systems can identify ``digital biomarkers'' that provide clinicians with objective, high-frequency measurements, effectively complementing traditional assessments with data-driven evidence.

Despite the promise of deep learning in this domain, its large-scale deployment is severely restricted by the sensitive nature of pediatric healthcare data. Under strict regulations such as the Health Insurance Portability and Accountability Act (HIPAA) in the US and the General Data Protection Regulation (GDPR) in the EU, raw RGB video recordings of children are considered primary biometric identifiers. These recordings contain sensitive Protected Health Information (PHI) and cannot be easily shared across institutions or aggregated in centralized cloud servers due to the risk of identity disclosure \cite{Sheller2020FederatedLI}. Furthermore, the Family Educational Rights and Privacy Act (FERPA) often applies when behavioral data is collected within educational settings, creating a complex multi-jurisdictional compliance burden for multi-site research. While skeleton-based representations have been proposed as a privacy-preserving alternative \cite{li2023mmasd}, motion-based biometric ``fingerprints'' still pose significant re-identification risks when large datasets are centralized. Consequently, clinical data typically remains isolated in ``data silos'', leading to small-scale studies that fail to capture the full spectrum of behavioral diversity across different demographics, age groups, and severity levels.

To overcome this obstacle, we propose a robust, privacy-preserving framework for pose-based autism behavior recognition using Federated Learning (FL) \cite{pmlr-v54-mcmahan17a}. Our approach shifts the training paradigm from the traditional ``data-to-model'' architecture to a decentralized ``model-to-data'' paradigm. This enables collaborative training across distributed clinical sites without ever transferring raw patient information or sensitive pose sequences. To ensure high recognition accuracy in this decentralized setting, we adopt \textbf{FreqMixFormer} \cite{wu2024frequency}, a state-of-the-art skeleton-based human action recognition approach. By leveraging frequency-domain analysis, FreqMixFormer can capture discriminative temporal motion patterns while maintaining low computational overhead, making it ideal for deployment on clinical edge nodes with varying hardware capabilities.

The core of our framework is a \textbf{two-layer privacy protection strategy}. First, we utilize 3D skeletal representations of human bodies rather than raw RGB video, filtering out identifiable facial features, clothing, and environmental contexts. Second, the federated optimization protocol ensures that even these abstracted pose sequences remain localized within each clinical site's secure infrastructure. Only anonymized model parameters are shared with a central aggregator, providing a double-layer defense against both visual and biometric identity disclosure while strictly aligning with the ``data minimization'' and ``privacy-by-design'' principles of modern healthcare regulations.

A critical challenge in this decentralized setting is the inherent \textit{heterogeneity} of multi-site clinical data. Variations in therapeutic protocols, room layouts, camera specifications, and patient demographics introduce significant domain shifts (Non-IID data) that standard federated algorithms often fail to resolve. To address this, we conduct a systematic investigation of various \textbf{Personalized Federated Learning (PFL)} strategies, including FedBN \cite{li2021fedbn}, FedPer \cite{arivazhagan2019federated}, and Adaptive Personalized Federated Learning (APFL) \cite{deng2020apfl}. Our framework evaluates how different personalization layers---ranging from local batch normalization to adaptive model mixing---can bridge the gap between global collaborative knowledge and site-specific behavioral nuances.

To validate our framework, we establish a simulated multi-site benchmark using the MMASD dataset \cite{li2023mmasd}. Through extensive experiments across multiple clinical themes, including Robotic-assisted therapy, Rhythm-based activities, and Yoga-based poses, we demonstrate that while standard aggregation methods provide a collaborative baseline, adaptive personalization is essential for handling severe clinical heterogeneity. Our results suggest that APFL, in particular, offers a superior balance of global representation and local adaptation, achieving performance comparable to centralized training while strictly maintaining data residency.

The primary contributions of this work are summarized as follows:
\begin{itemize}
    \item We present the first comprehensive framework for pose-based child autism behavior recognition that integrates a \textbf{two-layer privacy strategy}, combining skeletal abstraction with decentralized federated optimization to comply with HIPAA/GDPR standards.
    \item We utilize the efficient \textbf{FreqMixFormer} architecture, incorporating frequency-aware attention to ensure robust behavioral analysis across distributed clinical nodes.
    \item We provide a \textbf{systematic investigation of personalized federated learning strategies} (FedBN, FedPer, APFL) to address cross-site clinical heterogeneity and domain shift in autism clinical data.
    \item We establish a \textbf{federated benchmark protocol} on the MMASD benchmark, demonstrating that \textbf{adaptive personalization} via APFL provides a robust pathway for multi-site clinical AI without compromising patient anonymity.
\end{itemize}

\section{Related Work}
\label{sec:related}

\subsection{Pose-based Autism Behavior Recognition}
Automated assessment of Autism Spectrum Disorder (ASD) has transitioned from qualitative observations to quantitative metrics using advanced computer vision techniques. Early studies primarily focused on facial expression analysis and repetitive behavior detection (e.g., hand flapping, head banging) using raw RGB video \cite{perochon2023early, barami2024automated}. However, the use of raw video poses significant privacy risks for pediatric subjects, especially in rare disease contexts where background cues can lead to re-identification. To address this, skeleton-based action recognition has emerged as a robust, privacy-preserving alternative. Early graph-based methods such as ST-GCN \cite{yan2018spatial} and AS-GCN \cite{shi2020skeleton} established the foundation for modeling spatiotemporal joint dynamics by representing the human body as a structured graph. More recently, Shift-GCN \cite{Cheng_2020_CVPR} and MS-G3D \cite{Liu_2020_CVPR} have further improved recognition accuracy by introducing multi-scale aggregation and efficient shift operations.

These models have been adapted for ASD stimming behavior detection, capturing complex motion patterns without relying on identifiable appearance information \cite{shamhan2025advancements, said2024imitasd}. The introduction of the MMASD dataset \cite{li2023mmasd} and its enhanced version MMASD+ \cite{ravva2024mmasdnoveldatasetprivacypervserving+} has provided the community with multimodal benchmarks (including 3D skeletons, optical flow, and body mesh) specifically tailored for intervention analysis. Despite these advances, most existing models rely on centralized data collection, which is often prohibited in multi-site clinical research due to institutional data-sharing restrictions and regulatory hurdles.

\subsection{Federated Learning in Clinical Settings}
Federated Learning (FL) \cite{pmlr-v54-mcmahan17a} offers a privacy-preserving paradigm that allows multiple institutions to collaboratively train a global model while keeping their raw data localized. In the healthcare domain, FL has been widely adopted to facilitate multi-institutional collaborations in medical imaging and genomics without compromising patient privacy \cite{Sheller2020FederatedLI, xu2021federated}. For ASD specifically, Shamseddine et al. \cite{shamseddine2022feasibility} demonstrated the feasibility of FL for detecting autism using behavioral traits and facial features. Recent works have extended this decentralized approach to serious games \cite{pavlidis2024federated} and smart assistant systems for behavioral monitoring \cite{hegiste2025federated}.

A significant challenge in multi-site clinical FL is the risk of ``gradient conflict'', where divergent data distributions across clinics (e.g., variations in age groups or therapeutic activities) pull the global model in opposing directions. Advanced aggregation algorithms, such as FedAP \cite{lu2024personalized}, have been proposed to mitigate these conflicts by incorporating adaptive batch normalization or gradient coordination. Our work builds on these foundations by proposing a decentralized framework that utilizes skeletal representations to provide a double layer of privacy protection tailored for autism clinical research.

\subsection{Personalized Federated Learning}
The primary bottleneck in clinical FL is the heterogeneity of data distributions across sites, often referred to as the Non-IID (not identically and independently distributed) problem. Standard FL algorithms like FedAvg \cite{pmlr-v54-mcmahan17a} and the regularization-based FedProx \cite{li2020federated} often struggle to generalize when site-specific domain shifts are severe. Personalized Federated Learning (PFL) \cite{tan2022towards} addresses this by allowing sites to maintain local model variants that adapt the global representation to their unique local distribution. 

Seminal PFL approaches include Per-FedAvg \cite{fallah2020personalized}, which utilizes meta-learning for rapid local adaptation, and Ditto \cite{pmlr-v139-li21h}, which regularizes local models toward the global representation while allowing for significant local specialization. Structural personalization techniques, such as FedBN \cite{li2021fedbn} and FedPer \cite{arivazhagan2019federated}, focus on keeping specific layers (e.g., batch normalization or classifier heads) local to capture site-specific statistics. In this work, we specifically leverage the Adaptive Personalized Federated Learning (APFL) algorithm \cite{deng2020apfl}, which adaptively learns the optimal mixing ratio between global and local knowledge. This ensures that our autism behavior recognition model can capture universal behavioral markers while remaining sensitive to the unique phenotypic expressions encountered at individual clinical sites.

\section{Methodology}
\label{sec:method}

In this section, we detail our proposed framework for privacy-preserving autism behavior recognition. Our approach integrates a two-layer privacy protection mechanism that combines skeletal abstraction with decentralized optimization. To ensure both high recognition fidelity and computational efficiency, we adopt the \textbf{FreqMixFormer}~\cite{wu2024frequency} architecture as our core behavioral analysis backbone. Furthermore, we investigate several Personalized Federated Learning (PFL) strategies to handle the inherent heterogeneity in multi-site clinical data.

\begin{figure*}[t]
  \centering
  \includegraphics[width=0.85\linewidth]{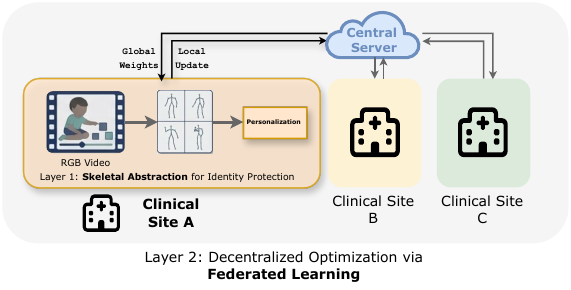}
  \vspace{-0.4cm}
   \caption{\textbf{Overview of the proposed Two-Layer Privacy framework.} The first layer achieves privacy via skeletal abstraction, filtering out raw biometric identifiers from videos. This is done at each clinical site for their local video data (a demonstration is shown for `Clinical Site A' in this figure). The second layer employs decentralized optimization (FL) using the efficient FreqMixFormer backbone to maintain data residency. Adaptive personalization layers (FedBN, FedPer, or APFL) are evaluated to handle clinical heterogeneity by adaptively mixing global knowledge with local specialization.}
   \label{fig:framework}
\end{figure*}

\subsection{Problem Formulation}
We consider a decentralized clinical network comprising $N$ participating sites (clients). Each site $i \in \{1, \dots, N\}$ maintains a private dataset $\mathcal{D}_i = \{(x_{i,j}, y_{i,j})\}_{j=1}^{n_i}$, where $x_{i,j}$ denotes an abstracted behavioral sequence and $y_{i,j} \in \{1, \dots, C\}$ is the corresponding clinical label. Due to the sensitive nature of pediatric data, raw sequences $x_{i,j}$ are restricted to local storage within the site's secure infrastructure. Our objective is to learn a set of personalized models $\{v_i\}_{i=1}^N$ that minimize the collective empirical risk:
\begin{equation}
    \min_{\{v_1, \dots, v_N\}} \sum_{i=1}^N \frac{n_i}{n} F_i(v_i), \quad F_i(v_i) = \frac{1}{n_i} \sum_{j=1}^{n_i} \ell(v_i; x_{i,j}, y_{i,j}),
\end{equation}
where $n = \sum n_i$ and $v_i$ represents the model deployed at site $i$. Depending on the personalization strategy, $v_i$ may be a site-specific variant of a shared global model $w$.

\subsection{Skeletal Abstraction for Identity Protection}
The primary challenge in analyzing pediatric autism behaviors lies in the trade-off between clinical utility and patient privacy. Raw RGB video recordings are highly identifiable, containing facial features, home environments, and other sensitive visual cues that constitute Protected Health Information (PHI). To mitigate this risk, our framework adopts 3D skeletal representations $S \in \mathbb{R}^{T \times V \times 3}$ as the foundational data modality, where $T$ denotes temporal length and $V$ denotes the number of joint keypoints.

By abstracting behavioral sequences into 3D joint trajectories, we decouple critical motion patterns from identifiable appearance information. This transformation serves as the \textbf{first layer of privacy}, ensuring that identifiable patient characteristics are filtered out before any model training or collaborative optimization occurs. Furthermore, skeletal data is naturally invariant to environmental factors such as illumination, background clutter, and camera-specific color distributions, which often vary across clinical rooms. In this work, we leverage 3D joint coordinates from the MMASD benchmark \cite{li2023mmasd}, which features 71 keypoints (including SMPL, extra, and H36M joints). A visualization is presented in Figure~\ref{fig:skeleton_samples}. Note that our framework remains compatible with other advanced world-grounded pose estimation models like GVHMR \cite{shen2024gvhmr} for enhanced temporal fidelity.

\subsection{Action Recognition Backbone: FreqMixFormer}
To perform robust behavior analysis on the extracted skeletons, we employ \textbf{FreqMixFormer} \cite{wu2024frequency} as our model backbone. Unlike traditional GCN-based methods that primarily operate in the spatial domain, FreqMixFormer utilizes a frequency-aware mixed transformer architecture that is specifically optimized for skeletal action recognition.

The model incorporates a frequency-aware attention module that leverages the Discrete Cosine Transform (DCT)~\cite{ahmed1974discrete} to capture discriminative motion patterns in the frequency domain. By analyzing the frequency components of joint trajectories, the framework effectively distinguishes between similar repetitive behaviors, such as subtle rhythmic hand movements versus stereotypical hand flapping. FreqMixFormer employs a mixed transformer architecture that balances global temporal dependencies with local spatial correlations while maintaining high computational and memory efficiency. This lightweight design is particularly advantageous for Federated Learning settings; by minimizing the parameter footprint, the model reduces the hardware overhead for on-device training at decentralized clinical nodes and significantly lowers the communication costs associated with transmitting model weights or gradients across multiple FL rounds.

\subsection{Decentralized Optimization via Federated Learning}
Even with skeletal abstraction, the accumulation of large-scale clinical data at a single institution remains a significant regulatory and ethical bottleneck. We therefore employ Federated Learning (FL) as the \textbf{second layer of privacy}, shifting the training paradigm from centralized aggregation to decentralized optimization. This approach ensures that even anonymized pose sequences never leave the secure infrastructure of the participating clinical site, maintaining strict data residency and subject privacy.

The optimization proceeds in a server-client architecture. In each communication round $t$, a central server broadcasts the current global FreqMixFormer parameters $w^t$. Each clinical site $i$ performs $K$ local updates using Stochastic Gradient Descent (SGD) on its private data $\mathcal{D}_i$. To further mitigate system heterogeneity and stabilize training, we evaluate both standard FedAvg \cite{pmlr-v54-mcmahan17a} and FedProx \cite{li2020federated} aggregation schemes. Only the resulting model updates $\Delta w_i^t$ are shared with the server, which aggregates them using a weighted averaging scheme:
\begin{equation}
    w^{t+1} = \sum_{i=1}^N \frac{n_i}{n} (w^t + \Delta w_i^t).
\end{equation}
This collaborative optimization enables individual sites to benefit from a global representation of autistic behaviors without compromising institutional ownership of sensitive data.

\subsection{Addressing Heterogeneity via Personalization}
Clinical data is characterized by significant non-IID properties arising from diverse camera configurations, therapeutic protocols, and patient demographics. To resolve these domain shifts, we evaluate three prominent personalization strategies:

\paragraph{FedBN: Local Batch Normalization.} FedBN \cite{li2021fedbn} addresses feature-level heterogeneity by maintaining Batch Normalization (BN) layers locally at each site. While convolutional and linear weights are aggregated globally to learn shared behavioral representations, the local BN layers capture site-specific statistics (mean and variance), effectively performing a form of internal domain adaptation to the local clinical environment.

\paragraph{FedPer: Personalization Layers.} FedPer \cite{arivazhagan2019federated} decomposes the model architecture into a shared backbone (base layers) and a local head (personalization layers). During the federated process, only the backbone parameters are aggregated, while each site optimizes its own classifier head tailored to its specific label distribution and behavioral nuances.

\paragraph{APFL: Adaptive Mixing.} APFL \cite{deng2020apfl} provides a more flexible approach by formulating the personalized model $v_i$ as a weighted mixture of the global model $w$ and a local-only model $u_i$: $v_i = \alpha_i u_i + (1 - \alpha_i) w$. The mixing parameter $\alpha_i \in [0, 1]$ is updated adaptively via gradient descent to minimize local validation loss:
\begin{equation}
    \alpha_i^{t+1} = \alpha_i^t - \eta_\alpha \langle \nabla f_i(v_i), u_i - w \rangle.
\end{equation}
This mechanism allows sites to automatically determine the optimal balance between global knowledge and local specialization. As demonstrated in our experiments, this adaptive strategy consistently provides the best performance in heterogeneous clinical environments.

\section{Experiments}
\label{sec:experiments}

In this section, we conduct a systematic evaluation of our framework to address two primary research questions. First, we investigate whether standard Federated Learning (FL) can effectively capture autistic behavioral patterns across heterogeneous sites (RQ1). Second, we evaluate various personalization strategies to determine the optimal balance between global knowledge and site-specific adaptation (RQ2).

\subsection{Experimental Setup}
\begin{figure}[t]
  \centering
  \begin{subfigure}{\linewidth}
    \centering
    \includegraphics[width=\linewidth]{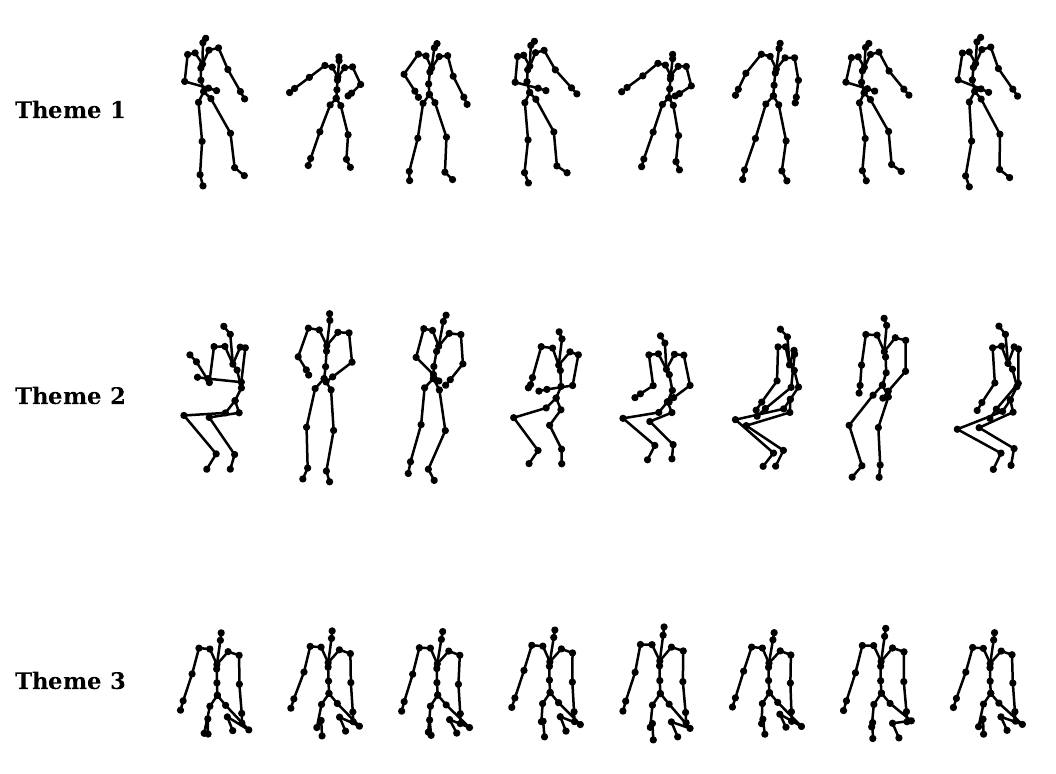}
    \caption{Temporal trajectories across different clinical themes.}
    \label{fig:skel_seq}
  \end{subfigure}
  \vspace{0.2cm}
  \begin{subfigure}{\linewidth}
    \centering
    \includegraphics[width=0.8\linewidth]{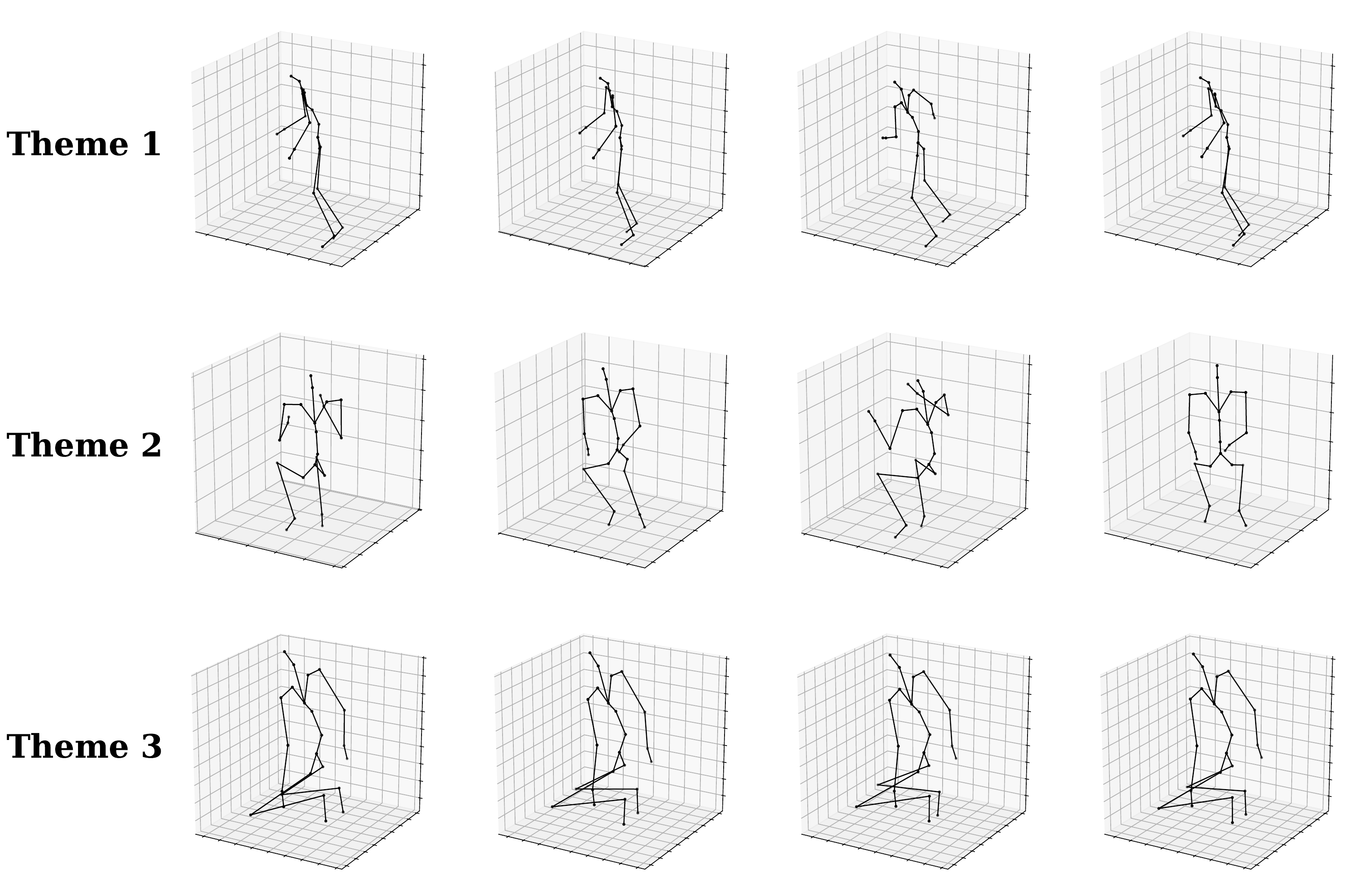}
    \caption{3D joint representation of an anonymized sample.}
    \label{fig:skel_3d}
  \end{subfigure}
   \caption{\textbf{Visualization of 3D skeletal data from the MMASD benchmark.} (a) shows representative sequences for Robotic-assisted therapy, Rhythm-based activities, and Yoga-based poses. (b) illustrates the 3D joint structure. This abstraction preserves kinetic motion for behavior analysis while raw biometric identifiers are removed.}
   \label{fig:skeleton_samples}
   \vspace{-3mm}
\end{figure}

\paragraph{Dataset: MMASD Benchmark.} We evaluate our framework using the MultiModal ASD (MMASD) benchmark \cite{li2023mmasd}, a privacy-preserving dataset specifically designed for autism intervention analysis. The dataset consists of 1,315 segmented samples derived from over 100 hours of intervention recordings involving 32 children diagnosed with ASD. To facilitate robust behavior analysis, MMASD provides four primary modalities; our framework specifically utilizes the \textbf{3D skeleton} data, which features 71 body keypoints extracted via the ROMP algorithm. Figure~\ref{fig:skeleton_samples} illustrates representative 3D skeletal samples from the benchmark, demonstrating how behavioral motion is captured while filtering out identifiable visual appearance information.

\paragraph{Clinical Themes and Multi-site Partitioning.} The behavioral samples in MMASD are categorized into three distinct clinical themes reflecting different therapeutic activities: (1) \textit{Robotic-assisted therapy}, comprising `Arm Swing', `Body Swing', `Chest Expansion', and `Squat'; (2) \textit{Rhythm-based activities}, including `Drumming', `Maracas Forward Shaking', `Maracas Shaking', and `Sing and Clap'; and (3) \textit{Yoga-based poses}, such as `Frog Pose', `Tree Pose', and `Twist Pose'. 

To simulate a distributed clinical learning environment, we adopt a \textbf{cross-silo Federated Learning} configuration. In cross-silo FL, a small number of trusted institutions (e.g., hospitals, clinics, or research centers) collaboratively train models while keeping patient data localized. Unlike cross-device FL, which typically involves thousands or millions of unpredictable edge devices, cross-silo settings usually consist of only a handful of participating sites with relatively larger datasets and stable computational resources. This setting is particularly relevant for clinical research, where collaboration typically occurs among a few verified medical institutions. Following this paradigm, we partition the dataset into \textit{three simulated clients}, each corresponding to one clinical theme. This design reflects realistic healthcare collaborations in which specialized institutions often focus on distinct therapeutic protocols or patient cohorts. Consequently, each client exhibits a unique data distribution, introducing meaningful domain shifts across sites. By treating these theme-specific cohorts as individual silos, our setup captures the key challenge of cross-silo federated learning: leveraging complementary knowledge across a small number of heterogeneous clinical institutions while preserving strict data locality.

\paragraph{Model Backbone.} All experiments utilize {FreqMixFormer} \cite{wu2024frequency} as the shared action recognition backbone trained from scratch. Its frequency-aware attention and efficient parameterization allow for high-fidelity behavior analysis even on clinical edge devices with limited resources.

\begin{table*}[h!]
\centering
\small
\caption{\textbf{Comparative performance summary across clinical themes.} We categorize methods into Baseline (Local), Standard FL, and Personalized FL (PFL). Accuracy results demonstrate that adaptive personalization (APFL) consistently outperforms traditional federated baselines and isolated local training.}
\label{tab:results}
\resizebox{0.65\linewidth}{!}{
\begin{tabular}{llcccc}
\toprule
 & Method & Theme 1 & Theme 2 & Theme 3 & Avg (\%) \\
\midrule
\textbf{Baseline} & Local  & 87.10 & 65.33 & 95.41 & 82.61 \\
\midrule
\textbf{FL} & FedAvg  & 70.16 & 52.67 & 88.07 & 70.30 \\
& FedProx  & 79.03 & 70.00 & \textbf{98.17} & 82.40 \\
\midrule
\textbf{Personalized FL} & FedBN  & 66.13 & 78.67 & 64.22 & 69.67 \\
& FedPer  & 63.71 & 74.67 & 91.74 & 76.71 \\
& APFL & \textbf{92.74} & \textbf{78.00} & 92.66 & \textbf{87.80} \\
\bottomrule
\end{tabular}
}
\end{table*}

\paragraph{Implementation Details.} We train our framework for a total of {30 communication rounds}. In each round, every client performs {1 local epoch} of training. We utilize Stochastic Gradient Descent (SGD) as the local optimizer with a learning rate of {0.1}, a weight decay of {$5 \times 10^{-4}$}, and a momentum of {0.9}. To ensure stable convergence in the early stages of decentralized optimization, we implement a {warmup phase of 5 rounds} for all federated and local baselines.

\paragraph{Baselines and Comparison Groups.} We categorize the evaluated optimization strategies into three groups:
\begin{itemize}
    \item \textbf{Baseline (Local):} A site-specific training scheme where models are trained independently using only local data. This represents the upper bound for purely local adaptation without collaborative knowledge.
    \item \textbf{Federated Learning (FL):} Standard decentralized optimization schemes, including FedAvg \cite{pmlr-v54-mcmahan17a} and the regularization-based FedProx \cite{li2020federated}.
    \item \textbf{Personalized Federated Learning (PFL):} Methods designed to bridge the gap between global and local knowledge, including parameter-wise isolation (FedBN \cite{li2021fedbn}, FedPer \cite{arivazhagan2019federated}) and our primary adaptive strategy, APFL \cite{deng2020apfl}.
\end{itemize}
For APFL, the mixing parameter is initialized at $\alpha = 0.01$ to provide a strong global prior.

\begin{figure*}[t]
  \centering
  \vspace{-2mm}
  \includegraphics[width=0.95\linewidth]{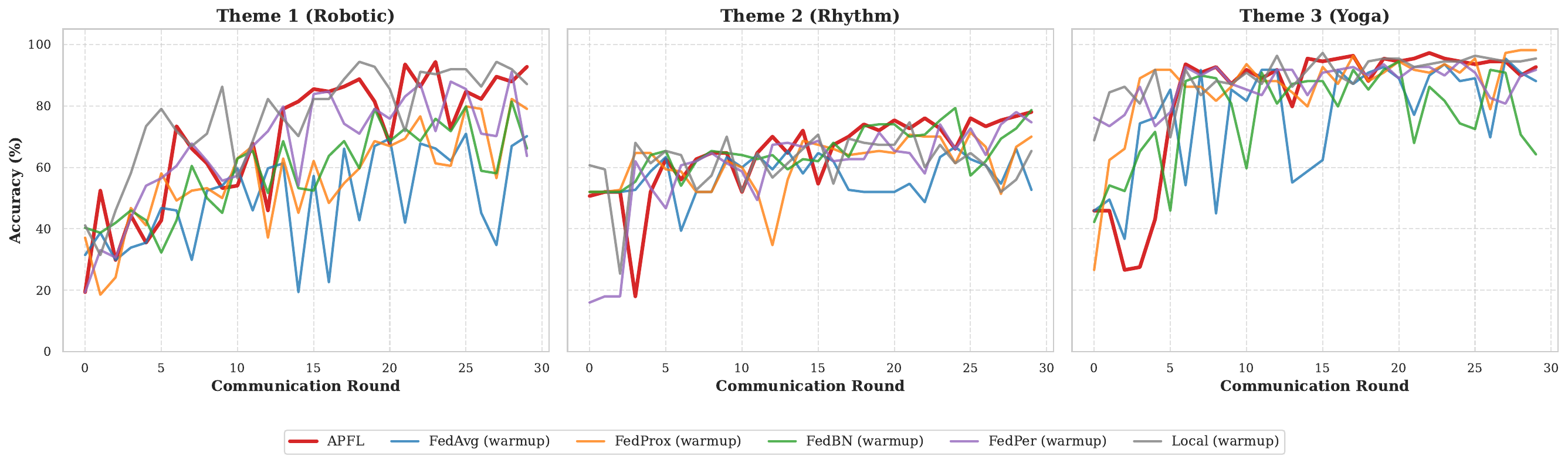}
   \caption{\textbf{Performance evolution across 30 communication rounds for different clinical themes.} The curves demonstrate the convergence characteristics of standard FL, parameter-wise PFL, and our adaptive personalization approach. APFL (red) consistently achieves superior stability and final recognition accuracy across all therapeutic domains.}
   \label{fig:accuracy_evolution}
\end{figure*}

\begin{figure*}[h!]
  \centering
  \includegraphics[width=0.95\linewidth]{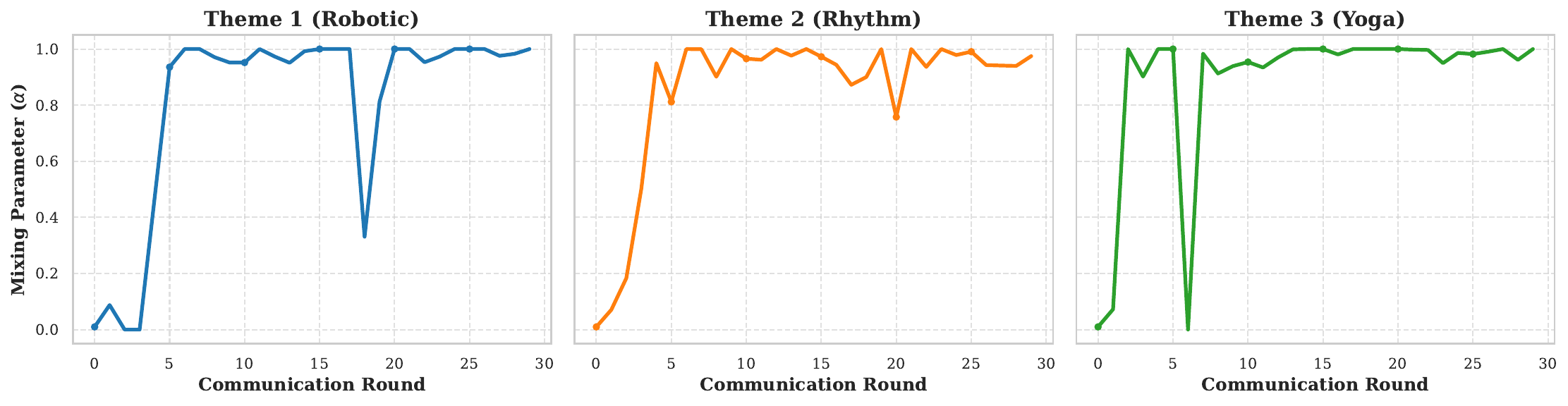}
   \caption{\textbf{Evolution of the adaptive mixing parameter $\alpha$ across different clinical themes.} The parameter is initialized with a low value and adaptively increases, indicating that the model initially prioritizes global knowledge synthesized from the collaborative network before gradually incorporating site-specific behavioral nuances from local data.}
   \label{fig:alpha_evolution}
\end{figure*}

\subsection{Performance Analysis}
\paragraph{Effectiveness of Federated Learning (RQ1).} Our comparative analysis reveals that traditional federated learning (FL) faces significant challenges in the clinical domain. As shown in Table~\ref{tab:results} and Figure~\ref{fig:accuracy_evolution}, standard FedAvg achieves an average accuracy of 70.30\%, significantly lagging behind the 82.61\% achieved by independent Local training. This gap is particularly severe in Theme 1, where FedAvg lags by nearly 17\%. While FedProx improves robustness, it only manages to align with the local baseline rather than surpassing it. These findings confirm that simple model aggregation is often counterproductive for autism behavior recognition, as site-specific domain shifts (Non-IID data) mask universal behavioral markers. The results suggest that for autism research, a monolithic global model is insufficient to capture the full spectrum of behavioral diversity across heterogeneous clinical sites.

\paragraph{Optimization via Personalized Federated Learning (RQ2).} To overcome clinical heterogeneity, we evaluate various PFL strategies. Parameter-wise isolation methods, such as FedBN (69.67\%) and FedPer (76.71\%), show limited efficacy. As illustrated in the convergence curves, these methods often exhibit unstable training dynamics and fail to consistently outperform the local baseline. This suggests that isolating specific layers like batch normalization or classifier heads is insufficient for capturing the complex spatiotemporal nuances of 3D skeletal data across diverse therapeutic themes.

The APFL algorithm emerges as the superior solution, achieving the highest average accuracy of 87.80\%, which represents a significant improvement over the 82.61\% average accuracy of the local baseline. While isolated local training provides a strong benchmark for site-specific performance, APFL consistently surpasses these results by adaptively learning the optimal mixing ratio between global representations and local specialization. This performance gain demonstrates the critical benefit of leveraging collective data knowledge from other clinical sites; by adaptively incorporating global information, APFL enables each clinical silo to benefit from behavioral markers learned across the entire network, effectively overcoming the limitations of purely local data. Unlike standard FL methods which may be degraded by clinical heterogeneity, our adaptive personalized approach recovers and enhances local performance, proving that collaboration across clinical silos is highly advantageous when appropriate personalization mechanisms are employed. Furthermore, APFL achieves a peak accuracy of 92.74\% in Theme 1 and maintains stable convergence, as shown in Figure~\ref{fig:accuracy_evolution}. This adaptive mechanism proves essential for multi-site clinical research, enabling the framework to overcome the limitations of both monolithic global models and isolated local training.

\section{Discussion}
\label{sec:discussion}

The experimental findings presented in Section~\ref{sec:experiments} underscore several critical challenges and opportunities in the field of medical federated learning, particularly for pediatric autism research. In this section, we provide a deeper analysis of the trade-off between cross-site generalization and local specialization, the adaptive dynamics of the APFL algorithm, and the broader regulatory implications of our two-layer privacy framework.

\paragraph{Generalization vs. Personalization in ASD Recognition}
A fundamental challenge in multi-site clinical AI is the tension between learning universal behavioral markers and adapting to local clinical nuances. The performance of our FreqMixFormer backbone under different optimization schemes suggests that autistic behaviors are highly context-dependent. The failure of traditional FedAvg to consistently outperform isolated local training indicates that monolithic ``one-size-fits-all'' models are ill-suited for the complex, non-IID distributions encountered in autism clinical data. This heterogeneity is likely driven by differences in clinical therapeutic protocols, individual patient support needs, and the high phenotypic variability inherent in Autism Spectrum Disorder.

Our observation that adaptive mixing (APFL) outperforms parameter-wise personalization methods (FedBN, FedPer) suggests that for high-dimensional temporal data like 3D skeletons, local adaptation must occur holistically. While methods like FedBN only adapt internal feature statistics, APFL allows the entire model representation to be adaptively weighted based on site-specific characteristics. This indicates that successful multi-site clinical AI requires frameworks that treat global collaborative knowledge as a robust prior that can be flexibly modulated by local behavioral evidence.

\paragraph{The Role of Adaptive Mixing in Clinical Heterogeneity}
To further investigate the personalization mechanism, we analyze the evolution of the adaptive mixing parameter $\alpha$ throughout the federated training process. As illustrated in Figure~\ref{fig:alpha_evolution}, $\alpha$ starts with a relatively low value and gradually increases across all clinical themes. This trajectory provides two key insights. First, the low initial value indicates that the model relies heavily on the global representation in the early stages, effectively using the collaborative network to establish a stable behavioral foundation. Second, the adaptive increase toward 1 demonstrates that as the local model matures, the system automatically shifts its focus toward site-specific behavioral nuances. This confirms that the global model acts as a critical knowledge anchor that helps stabilize local learning, particularly for sites with limited data samples. The ability to automatically tune this balance makes APFL particularly robust for real-world clinical deployment, where the optimal degree of personalization may vary significantly across institutions.

\paragraph{Regulatory Implications and Visual Privacy}
Beyond performance metrics, the success of our Two-Layer Privacy framework has significant implications for the future of clinical collaboration. By combining skeletal abstraction with decentralized optimization, we provide a mathematically and regulatorily sound pathway for building large-scale behavioral models.

The first layer of privacy—skeletal abstraction—addresses the primary biometric identifier (the human face) while the second layer—Federated Learning—addresses the administrative bottleneck of data residency. This dual-layer approach aligns strictly with HIPAA's ``privacy-by-design'' and GDPR's ``data minimization'' principles. By ensuring that raw identifiers are filtered at the source and that abstracted motion sequences never leave the secure clinical infrastructure, we address the ``data silo'' problem that has historically impeded progress in autism research. This framework establishes a standard for multi-site collaboration that respects patient autonomy while enabling the development of more powerful, objective diagnostic tools. Moving forward, the integration of such privacy-preserving technologies will be essential for democratizing access to high-quality clinical AI across diverse and under-resourced healthcare settings.

\section{Conclusion}
\label{sec:conclusion}
In this work, we introduced a robust, privacy-first framework for pose-based child autism behavior recognition. By integrating a \textbf{two-layer privacy protection strategy}---utilizing skeletal abstraction and cross-silo federated learning---we provide a secure pathway for collaboration among a trusted network of clinical institutions. Our systematic evaluation of the efficient \textbf{FreqMixFormer} architecture across various federated optimization schemes highlights the critical role of personalization in handling clinical data heterogeneity. We demonstrate that while traditional FL often struggles with non-IID behavioral data, adaptive personalization via \textbf{APFL} significantly outperforms isolated local training, proving the substantial benefit of leveraging shared knowledge across clinical sites while strictly maintaining data residency.
Future research will explore the integration of \textit{multimodal data streams} to further enhance diagnostic precision and behavioral insight. Specifically, incorporating \textit{speech patterns}, \textit{vocal prosody}, and the \textit{conversational dynamics} between patients and clinicians or behavior analysts could provide deeper insights into social-communication nuances that are currently underserved by motion-only analysis. Furthermore, we aim to investigate the use of privacy-preserving large language models (LLMs) to analyze clinical notes and transcripts in a federated manner. Expanding the framework to handle such heterogeneous, high-dimensional data streams while maintaining rigorous privacy standards for sensitive audio-visual interactions remains a promising direction for the next generation of automated autism intervention and monitoring tools.

{
    \small
    \bibliographystyle{ieeenat_fullname}
    \bibliography{main}
}


\end{document}